# Learning on tree architectures outperforms a convolutional feedforward network

Yuval Meir[1], Itamar Ben-Noam[1], Yarden Tzach[1], Shiri Hodassman[1] and Ido Kanter[1,2*]

[1]Department of Physics, Bar-Ilan University, Ramat-Gan, 52900, Israel.

[2]Gonda Interdisciplinary Brain Research Center, Bar-Ilan University, Ramat-Gan, 52900, Israel.

[*]Corresponding author email: ido.kanter@biu.ac.il

## ABSTRACT

Advanced deep learning architectures consist of tens of fully connected and convolutional hidden layers, currently extended to hundreds, are far from their biological realization. Their implausible biological dynamics relies on changing a weight in a non-local manner, as the number of routes between an output unit and a weight is typically large, using the backpropagation technique. Here, a 3-layer tree architecture inspired by experimental-based dendritic tree adaptations is developed and applied to the offline and online learning of the CIFAR-10 database. The proposed architecture outperforms the achievable success rates of the 5-layer convolutional LeNet. Moreover, the highly pruned tree backpropagation approach of the proposed architecture, where a single route connects an output unit and a weight, represents an efficient dendritic deep learning.

## Introduction

Traditionally, deep learning (DL) stems from human brain dynamics, where the connection strength between two neurons is modified following their relative activities via synaptic plasticity[1,2]. However, these two learning scenarios are essentially different[3]. First, efficient DL architectures require tens of feedforward hidden layers[4,5], currently extended to hundreds[6,7], whereas brain dynamics consist of only a few feedforward layers[8-10].

Second, DL architectures typically consist of many hidden layers that are mostly convolutional layers. The response of each convolutional layer is sensitive to the presence of a particular pattern or symmetry in limited areas of the input, where its repeated operations in the subsequent hidden layers are expected to reveal large-scale features characterizing a class of inputs[11-15]. Similar operations have been observed in the visual cortex; however, approximated convolutional wirings have been confirmed mainly from the retinal input to the first hidden layer[8,16].

The final complex aspect is the implausible biological realization of the backpropagation (BP) technique[17,18], an essential component in the current implementation of DL, which changes a weight in a non-local manner. In a supervised learning scenario, an input is presented to the feedforward network, and the distance between the current and desired output is computed using a given error function. Subsequently, each weight is updated to minimize the error function using the following BP procedure. Each route between the weight and an output unit contributes to its modification, where all components along this route, weights and nonlinear nodal activation functions, are combined. Finally, weights belonging to a convolutional layer are made equal to establish a filter. The number of routes between an output unit and a weight is typically large (Fig. 1a-b); thus, this enormous transportation of precise weight information can be performed effectively using fast vector and matrix operations implemented by parallel GPUs. However, this process is evidently beyond biological realization. Note that convolutional and fully connected layers are the main contributors to this enormous parallel transportation, excluding an adjacent convolution to the input layer and a fully connected layer adjacent to the output units[19].

This study presents a learning approach based on tree architectures, where each weight is connected to an output unit via only a single route (Fig. 1c-d). This type of tree architectures represents a step toward a plausible biological learning realization[19] and is inspired by recent experimental evidence on dendritic[20] and sub-dendritic adaptations[19] and their nonlinear amplification[21-23]. In particular, the timescales of dendritic adaptation depend on the training frequency and can be reduced to several seconds only[24]. The question at the center of our work is whether learning on a tree architecture, inspired by dendritic tree learning, can achieve success rates comparable to those obtained from more structured architectures consisting of several fully connected and convolutional layers. Here, the success rates of the proposed Tree-3 architecture, with only three hidden layers, are demonstrated to outperform the achievable LeNet-5 success rates on the CIFAR-10 database[25]. The LeNet-5 architecture (Fig. 1a) consists of five hidden layers, two convolutional layers, and three fully connected layers, such that a weight can be connected to an output through a large number of routes, which can exceed a million (Fig. 1b).

## Methods

**The Tree-3 architecture and initial weights.** The Tree-3 architecture (Fig. 1c) consists of M=16 branches. The first layer of each branch consists of K (6 or 15) filters of size (5×5) for each one of the three RGB channels. Each channel is convolved with its own set of K filters, resulting in 3×K different filters. The convolutional layer filters are identical among the M branches. The first layer terminates with a max-pooling consisting of (2×2) non-overlapped squares. The second layer consists of a tree sampling. For the CIFAR-10 dataset, this layer connects hidden units from the first layer using non-overlapping rectangles of size (2×2×7), two consecutive rows of the 14x14 square, without shared-weights, but with summation over the depth of K filters, yielding an output of (16×3×7) hidden units. The third layer fully connects the (16×3×7) hidden units of the second layer with the 10 output units, representing the 10 different labels. For the MNIST dataset, the input is (28×28) and after the (5×5) convolution the output of each filter is (24×24) which terminates as (12×12) hidden units after performing the max-pooling. The tree sampling layer connects hidden units from the first layer using non-overlapping rectangles of size (4×4×3), without shared-weights, but with summation over the depth of K filters, resulting in a layer of (16×3) hidden units. The third layer fully connects the (16×3) hidden units of the second layer with the 10 output units, representing the 10 different labels. For online learning, the ReLU activation function is used, whereas Sigmoid is used for offline learning, except for K = 15, M = 80, where the ReLU activation function is used. All weights are initialized using a Gaussian distribution with zero mean and standard deviation according to He normal initialization[26].

Details of the weight, input, and output sizes for each layer of the Tree-3 architecture are summarized below.

| Type | Weight size | Input size | Output size |
|---|---|---|---|
| Conv2d | 3 x K x 5 x 5 groups = 3 | 3 x 32 x 32 | 3K x 28 x 28 |
| MaxPool2d | 2 x 2 | 3K x 28 x 28 | 3K x 14 x 14 |
| Tree Sampling | 3K x M x 14 x 14 | 3K x 14 x 14 | 3M x 7 |
| FC | 21M x 10 | 3M x 7 | 10 |

**Data preprocessing.** Each input pixel of an image is divided by the maximal value for a pixel, 255, and next multiplied by 2 and subtracted by 1, such that its range is [-1, 1]. The performance was enhanced by using simple data augmentation derived from original images, such as flipping and translation of up to two pixels for each direction. For offline learning with K = 15 and M = 80, the translation was up to four pixels for each direction.

**Optimization.** The cross-entropy cost function was selected for the classification task and was minimized using the stochastic gradient descent algorithm. The maximal accuracy was determined by searching over the hyper-parameters, i.e., learning rate, momentum constant and weight decay. Cross validation was confirmed using several validation databases each consisting of 10,000 random examples as in the test set. The averaged results were within the standard deviation (Std) of the reported average success rates. Nesterov momentum[27] and L2 regularization method[28] were used.

**Number of paths – LeNet-5.** The number of different routes between a weight emerging from the input image to the first hidden layer and a single output unit is calculated as follows (Fig. 1b). Consider an output-hidden unit of the first hidden layer, belonging to one of the (14×14) output hidden units of a filter at a given branch. This hidden unit contributes to a maximum of 25 different convolutional operations for each filter at the second convolutional layer. The output of this layer results in 16×25 different routes. The max-pooling of the second layer reduces the number of different routes to 16×25/4 =100. Each of these routes splits to 120 in the third fully connected layer and splits again to 84 in the fourth fully connected layer. Hence, the total number of routes is 100×120×84 = 1,008,000 different routes.

**Hyper-parameters for offline learning (Table 1, upper panel).** The hyper-parameters η (learning rate), μ (momentum constant[27]) and α (regularization L2[28]), were optimized for offline learning with 200 epochs. For LeNet-5, using mini-batch size of 100, η = 0.1, μ = 0.9 and α = 1e-4. For Tree-3 (K = 6 or 15, M = 16 or M = 80), using mini-batch size of 100, η = 0.075, μ = 0.965 and α = 5e-5 and for 10 Tree-3 (K = 15, M = 80) architectures where each one has one output only, using mini-batch size of 100, η = 0.05, μ = 0.97 and α = 5e-5. The learning-rate scheduler for LeNet-5, η = 0.01, 0.005, 0.001 for epochs = [0, 100), [100, 150), [150, 200], respectively. For Tree-3 (K = 6, M = 16) η=0.075, 0.05, 0.01, 0.005, 0.001, 0.0001) for epochs = [0, 50), [50, 70), [70, 100), [100, 150), [150, 175), [175,200], respectively. For Tree-3 (K = 15, M = 16) η = 0.075, 0.05, 0.01, 0.0075, 0.003 for epochs = [0, 50), [50, 70), [70, 100), [100, 150), [150,200], respectively. For Tree-3 (K = 15, M = 80) and 10 Tree-3 (K = 15, M = 80), η decays by a factor of 0.6 every 20 epochs. For Tree-3, the weight decay constant changes after epoch 50 to 1e-5. For the MNIST dataset the optimized hyper-parameters were a mini-batch size of 100, η = 0.1, μ = 0.9 and α = 5e-4. The learning rate scheduler was the same as for Tree-3 (K = 15, M = 16), on the CIFAR-10 dataset.

**Hyper-parameters for online learning (Table 1, bottom panel).** The hyper-parameters mini-batch size, η (learning rate), μ (momentum constant[27]) and α (regularization L2[28]), were optimized for online learning using the following three different dataset sizes (50k, 25k, 12.5k) examples. For LeNet-5, using mini-batch sizes of (100, 100, 50), η = (0.012,

0.017, 0.012), μ = (0.96, 0.96, 0.94) and α = (1e-4, 3e-3, 8e-3), respectively. For Tree-3 (K = 6, M = 16), using mini-batch sizes of (100, 100, 50), η = (0.02, 0.03, 0.02), μ = (0.965, 0.965, 0.965) and α = (5e-7, 5e-6, 5e-5), respectively.

**Ten Tree-3 architectures**. Each Tree-3 architecture has only one output unit representing a class. The ten architectures have a common convolution layer and are trained in parallel, where eventually the softmax function is applied on the output of the ten different architectures.

**Pruned BP.** The gradient of a weight emerging from an input unit connected an output via a single route (Tree-3 architecture, Fig. 1c), with non-zero ReLU activation function, is given by $\Delta(W^{Conv}) = Input \cdot W^{Tree} \cdot W^{FC} \cdot (Output - Output_{desired})$, otherwise, its value is equal to zero.

**Statistics.** Statistics of the average success rates and their standard deviations for online and offline learning simulations were obtained using 20 samples. The statistics of the percentage of zero gradients and their standard deviations in Fig. 2 were obtained using 10 different samples each trained over 200 epochs.

**Hardware and Software.** We used Google Colab Pro and its available GPUs. We used Pytorch for all the programming processes.

## Results

The Tree-3 architecture consists of M parallel branches, each receiving the same (32×32) RGB inputs (Fig. 1c). Each branch is connected to a convolutional layer, consisting of K (5×5) filters for each of the three RGB input colors followed by (2×2) max-pooling, resulting in a (14×14) output layer for each filter in each branch. The 3×K filters are identical among the M branches. After a non-overlapping (2×2×7) sampling across K filters belonging to the same branch, a tree sampling, each branch leads to 21 (3×7) outputs. Hence, the second layer consists of 21×M outputs of the M branches that are now fully connected to the 10 output units, representing the 10 possible labels. Thus, each weight in Tree-3 is connected to an output unit via a single route only (Fig. 1d).

The asymptotic success rates after many epochs using the entire dataset, including data augmentation, for Tree-3 using the BP algorithm (Methods) are presented in Table 1 (upper panel), along with the success rates for LeNet-5. These results indicate that the success rate of Tree-3 outperforms that of LeNet-5. The parallel tree branches (Fig. 1c) successfully replace the second convolutional layer in LeNet-5 as well as the several fully connected layers (Fig. 1a).

For M = 16, the success rates are improved when K is increased from 6 to 15 (Table 1, upper panel). In addition, for K = 15, success rates are improved when M is increased from 16 to 80 (Table 1, upper panel), where M = 80 is selected according to the following argument. In deep learning algorithms, the product of the size of the feature maps and their amount remains constant along most of the convolutional layers. For instance, in VGG-Net[29], when the size of the feature maps shrinks from 112×112 to 56×56, the number of filters accordingly increases from 128 to 256. Similarly, in LeNet-5, when the size of the feature maps shrinks from 14×14 in the first convolutional layer to 5x5 in the second layer (Fig. 1a), the number of filters increases from 6 to 16 ($(14/5) \times (6/16) \simeq 1$). For the Tree-3 architecture (Fig. 1c), the size of the feature maps shrinks from 14×14 to 7×1 (which can be approximated as $\sqrt{7} \times \sqrt{7}$); hence, the number of M branches (representing the amount of filters) has to increase to $15 \times (14/\sqrt{7}) \simeq 80$. Indeed, the success rate increases from ~0.765 for M = 16 to ~0.79 for M = 80 (Table 1, upper panel). The optimization of the success rates on the M and K grid deserves further research.

The Tree-3 architecture (Fig. 1c) deviates from a biological dendritic morphology. Although each weight, a dendritic segment, is connected to each output unit via one route, the architecture consists of 10 output units. We repeated our simulations for 10 Tree-3 architectures, where each one has only one output, and with the same cost function for the 10 output units (Methods). The results indicate that for K = 15 and M = 80, the success rates are improved to ~0.815.

In comparison to LeNet-5, the advantage of Tree-3 is enhanced when success rates were optimized for the online learning scenario, where each input was trained only once. Optimized success rates for online learning with 50K (the entire dataset), 25K, and 12.5K training inputs indicate that Tree-3 outperformed LeNet-5 by 6–8% (Table 1, lower panel).

Preliminary results indicate that the reported success rates of Tree-3 on CIFAR-10 can be further enhanced by a few percentage using the following two approaches. The first is adding weight crosses[19] to the 25 weights composed of the 5x5 convolutional filters (Fig. 1c). The second is to include an additional convolutional layer with a 1×1 filter[30] after the first convolution layer. This 1×1 convolutional layer preserves the tree architecture because it functions as a weighted sum of previous filters.

A similar comparison between the success rates of Tree-3 and LeNet-5 (Fig. 1) was also performed for the MNIST database[31]. The results indicate that the success rate of Tree-3 with many epochs is comparable to or even slightly better than that of LeNet-5. The reported success rate of LeNet-5 without augmentation was 0.9905[17], whereas Tree-3 obtained 0.9907 (Methods), which can be further enhanced using an additional 1×1 convolutional layer. This Tree-3 results on MNIST are significantly better than the recently achievable success rates for tree architectures without a convolution layer adjacent to the input[19].

## Discussion

An efficient approach for learning on tree architectures was demonstrated, where each weight is connected to an output unit via a single route only. The gradient, Δ, of a weight emerging from an input unit, $W^{Conv}$ (Fig. 2a), is given in the BP procedure as follows:

$$\Delta(W^{Conv}) = Input \cdot \sigma'_{Conv} \cdot W^{Tree} \cdot \sigma'_{Tree} \cdot W^{FC} \cdot (Output - Output_{desired})$$

where $\sigma'_{Conv}$ and $\sigma'_{Tree}$ are the derivatives of the current activation function of the hidden units in the convolutional and tree layers along the route, respectively (Fig. 2a). The last term denotes the difference between the current output and the desired one. When ReLU activation function is used, the gradient is simplified as follows:

$$\Delta(W^{Conv}) = Input \cdot W^{Tree} \cdot W^{FC} \cdot (Output - Output_{desired})$$

otherwise, the gradient is equal to zero. The simulations performed over each presented input in the test set indicate that the average fraction of zero gradients in the BP procedure was approximately 0.97, 0.88, and 0.72 for $W^{Conv}, W^{Tree}$, and $W^{FC}$, respectively (Fig. 2a-b). Since almost all possible gradients belong to $W^{Conv}$, the fraction of zero gradients in an entire BP step is ~0.97 (Methods). The simulations also indicate that these dominated zero gradient fractions (Fig. 2b) hold since the initial learning of a few hundred of presented inputs within the first epoch. These results indicate that dynamically, the number of necessary weight updates is minimized because the tree is being significantly pruned, which also reduces the complexity and the energy consumption of the learning algorithm. Noted that the strength distribution of the absolute values of the remaining ~0.03 fraction of non-zero gradients is wide and dominated by small values, likewise with the distribution of the relative weight changes, $|\Delta/W|$ (Supplementary Figure S1). These small changes might be dynamically neglected in each BP step without affecting the success rates, such that the required fraction of gradients might be minimized further, which may also be suitable to the biological reality of a limited signal-to-noise ratio. Preliminary results indicate that similar behavior occurs for the Sigmoid activation function, where large fractions of the absolute gradient values or relative weight changes are several order of magnitudes smaller than the dominating ones (Supplementary Figure S2). This indicates that the simplified utilization of BP for a high pruning tree might also apply to this scenario. From initial tests, it has been observed that dynamically neglecting absolute value gradients below a threshold, resulting in only 0.6% active gradients, barely affects the success rates.

In the Tree-3 architecture with K = 6 and M = 16, the maximal required number of calculated gradients in the convolutional layer, dominating the entire number of gradients, is ~5.7 million (5×5×3×6×28×28×16) (Fig. 1c). In the pruning tree BP, only ~0.6% is actually updated; hence, the remaining necessary updates are ~34,000, where the calculation of each one consists of ~2 operations only (see the abovementioned equation). In contrast, for LeNet-5, the maximal number of calculated gradients in the first convolutional layer is ~352,000 (5×5×3×6×28×28), where the max-pooling layer reduces the number to ~88,000. However, because each weight is connected to an output unit via an enormous number of routes (Fig. 1b and Methods), the computational complexity of each gradient is large. Hence, in principle, the computational complexity of LeNet-5 is significantly greater than that of the Tree-3 architecture with similar success rates; however, its efficient realization requires a new type of hardware.

Tree architectures represent a step toward a plausible biological realization of efficient dendritic learning and its powerful computation, where DL can be realized by highly pruned dendritic trees of a single or several neurons. The usefulness of a single convolutional layer adjacent to the input layer in preserving the tree architecture is evident. It enhances the success rates compared to tree architectures without convolution[19] and is consistent with the observed approximated convolutional wirings adjacent to the retinal input layer[8,16]. The efficient implementation of the BP technique on tree architectures, specifically for the dynamic of a highly pruning tree, requires a new type of hardware. Such hardware must be different from GPUs, which are better fitted to the current DL implementations, consisting of many convolutional and fully connected layers, based on vector and matrix multiplications.

Tree architecture learning is also expected to minimize the probability of occurrence and the effect of vanishing and exploding gradients, one of the challenges in DL[32]. In conventional DL architectures, such as that in Fig. 1a, each weight is updated via the summation of a large number of gradients associated with all its routes connecting the output units (Fig. 1b). Even if the gradient of each route is small, the summation can be large and lead to weight explosion, a reality which is not considered in the tree architectures. This possible advantage of shallow tree architectures in minimizing the effect of vanishing or exploding gradients, calls for their quantitative examination in further research.

The introduction of parallel branches instead of the second convolutional layer in LeNet-5 improved the success rates while preserving the tree structure. Each one of the M branches has the same convolutional filters emerging from the input layer. Nevertheless, as their weights in the second and the third layers differ, they have different effects on the output units (Fig. 1c). The first convolutional layer reveals feature maps in the inputs, similar to the function of the first convolutional layer in LeNet-5. Higher-order correlations among these feature maps are expected to be expressed via the summation of the depth of K different filters belonging to each branch, similar to the second convolutional layer in LeNet-5[13-15]. Preliminary simulations indicate that extending the tree architecture to the case where each branch can have different convolutional filters results in similar success rates. Hence, equalizing the filters among the branches after each BP step is unnecessary. The possibility that large-scale and deeper tree architectures, with an extended number of branches and filters, can compete with state-of-the-art CIFAR-10 success rates deserves further research. The first step was exemplified regarding to LeNet-5 and demonstrated the advantage of the dendritic learning concept and its powerful computation.

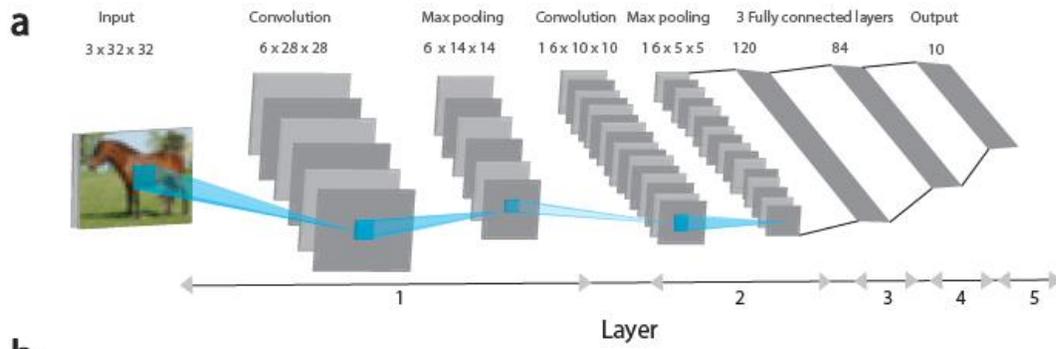
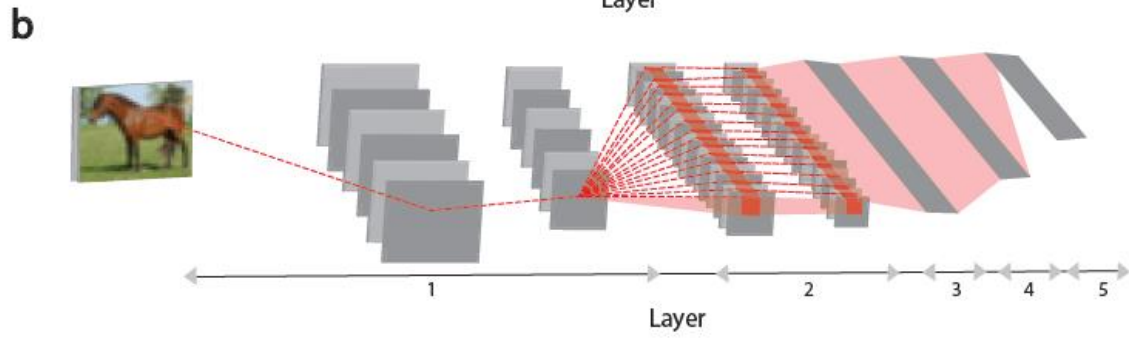
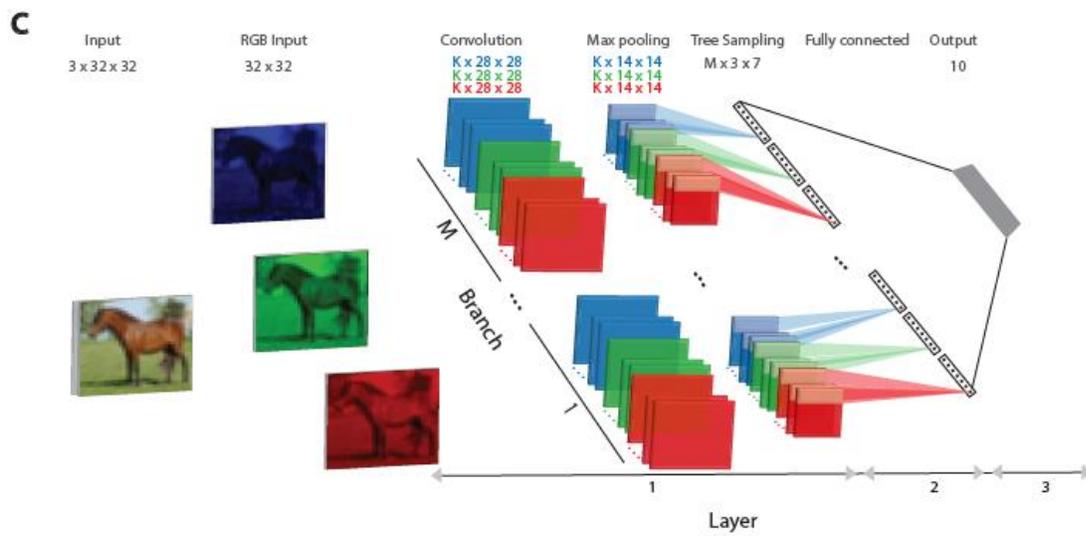
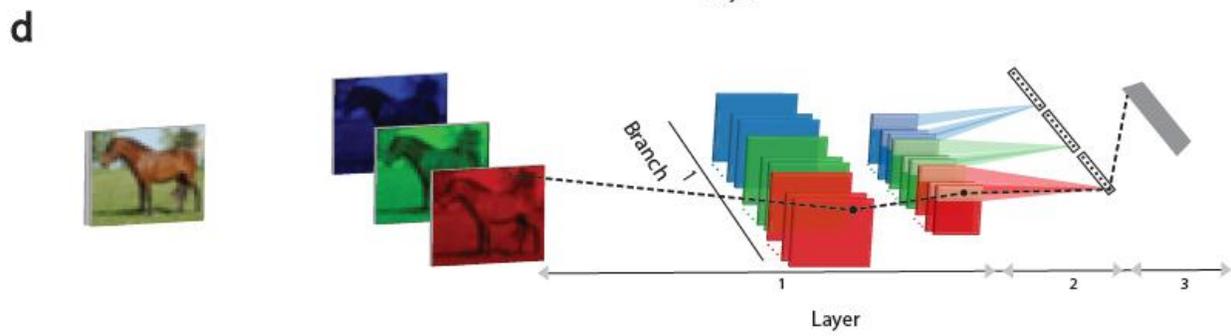

**Figure 1.** Examined convolutional LeNet-5 and Tree-3 architectures. (**a**) The convolutional LeNet-5 for the CIFAR-10 database consists of 32×32 RGB input images belonging to the 10 output labels. Layer 1 consists of six (5×5) convolutional filters followed by (2×2) max-pooling, Layer 2 consists of 16 (5×5) convolutional filters and Layers 3-5 have three fully connected hidden layers of sizes 400, 120, and 84, which are finally connected to the 10 output units. (**b**) Scheme of the routes affecting the updating of a weight belonging to Layer 1 in panel **a** (dashed red line) during the BP procedure. The weight is connected to one of the output units via multiple routes (dashed red lines) and can exceed one million (Methods). Note that all weights in Layer 1 are equalized to 6×(5×5) weights belonging to the six convolutional filters. (**c**) Examined Tree-3 structure consisting of M branches with the same 32×32 RGB input images. Layer 1 consists of 3×K (5×5) convolutional filters and K filters for each of the RGB input images. Each branch consists of the same 3×K filters, followed by (2×2) max-pooling which results in (14×14) output units for each filter. Layer 2 consists of a tree (non-overlapping) sampling (2×2×7 units) across the K filters for each RGB color in each branch, resulting in 21 (7×3) outputs for each branch. Layer 3 fully connects the 21×M outputs of the M branches of Layer 2 to the 10 output units. (**d**) Scheme of a single route (dashed black line) connecting an updated weight in Layer 1 (in panel **c**), during the BP procedure, to an output unit.

| Offline learning using many epochs - CIFAR-10 | | |
| --- | --- | --- |
| Model | Avg. Success rate | Std |
| LeNet-5 | 0.7535 | 0.0055 |
| Tree-3 (K = 6, M = 16) | 0.7502 | 0.0032 |
| Tree-3 (K = 15, M = 16) | 0.7670 | 0.0041 |
| Tree-3 (K = 15, M = 80) | 0.7913 | 0.0022 |

| Online learning using limited dataset - CIFAR-10 | | | | |
| --- | --- | --- | --- | --- |
| | LeNet-5 | | Tree-3 (K = 15, M = 16) | |
| No. examples | Avg. Success rate | Std | Avg. Success rate | Std |
| 50K | 0.5286 | 0.0131 | 0.6051 | 0.0046 |
| 25K | 0.4844 | 0.0124 | 0.5550 | 0.0092 |
| 12.5K | 0.4428 | 0.0098 | 0.5018 | 0.0083 |

**Table 1.** Comparison of offline and online learning success rates between LeNet-5 and Tree-3 architectures. Upper panel: Optimized success rates and their standard deviations for offline learning of LeNet-5 (first row), Tree-3 architecture with K=6 and M=16 (second row), with K = 15 and M = 16 (third row) and with K = 15 and M = 80 (fourth row); for detailed parameters, see Methods. Lower panel: Optimized success rates and their standard deviations for online learning, where each input is trained only once, for LeNet-5 and Tree-3 with K = 15 and M = 16, for 50K, 25K and 12.5K training inputs (Methods). Success rates of Tree-3 outperform those of LeNet-5 by 6–8%.

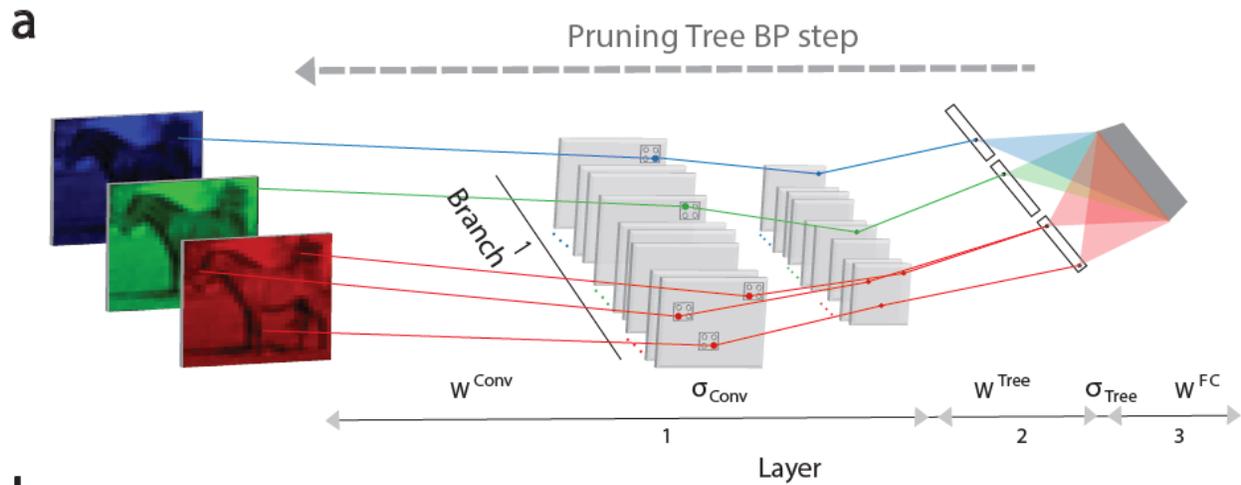

| Tree-3 (K = 15, M = 16) using many epoches - CIFAR-10 | | | |
|---|---|---|---|
| | Layer 1 | Layer 2 | Layer 3 |
| Avg. fraction zero gradients | 0.97 | 0.88 | 0.73 |

**Figure 2.** BP step on highly pruning Tree-3 architecture. (**a**) Scheme of a BP step in the first branch of a highly pruning Tree-3 architecture (Fig. 1d). The gray squares in the first layer represent convolutional hidden units, $\sigma_{Conv}$, and max-pooling hidden units that are equal zero, except several denoted by RGB dots. The non-zero tree output hidden units, $\sigma_{Tree}$, are denoted by black dots. The updated weights with nonzero gradients, in first layer, $W^{Conv}$, second layer, $W^{Tree}$, and third fully connected layer, $W^{FC}$, are denoted by RGB lines. (**b**) Fraction of zero gradients, averaged over the test set, and their standard deviations for the tree layers of Tree-3 architecture (K = 15, M = 16), after many epochs (Methods).

**Data availability**

Source data are provided with this paper. All data supporting the plots within this paper along with other findings of this study are available from the corresponding author upon reasonable request.

**Code availability**

The simulation code is provided with this paper, parallel to its publication, in GitHub.

**Acknowledgments**

I.K. acknowledges partial financial support of the Israel Ministry Science and Technology, via the grant, Brain-inspired ultra-fast and ultra-sharp machines for AI-assisted healthcare, via collaboration between Italy and Israel. S.H. acknowledges the support of the Israel Ministry Science and Technology.

**Author contributions**

Y.M. conducted all simulations together with I.B.N. and Y.T., where S.H. contributed to the simulations on MNIST and preparation of the figures. I.K. initiated the theoretical study of the examined tree architectures and supervised all aspects of the work. All authors commented on the manuscript.

**Competing interests**

The authors declare no competing interests.

**Additional information**
**Supplementary information** The online version contains supplementary material available at

# Learning on tree architectures outperforms a convolutional feedforward network

**Yuval Meir[1], Itamar Ben-Noam[1], Yarden Tzach[1], Shiri Hodassman[1] and Ido Kanter[1,2*]**

[1]Department of Physics, Bar-Ilan University, Ramat-Gan, 52900, Israel.

[2]Gonda Interdisciplinary Brain Research Center, Bar-Ilan University, Ramat-Gan, 52900, Israel.

[*]Corresponding author email: ido.kanter@biu.ac.il

**This PDF file includes:**

Figures. S1 and S2

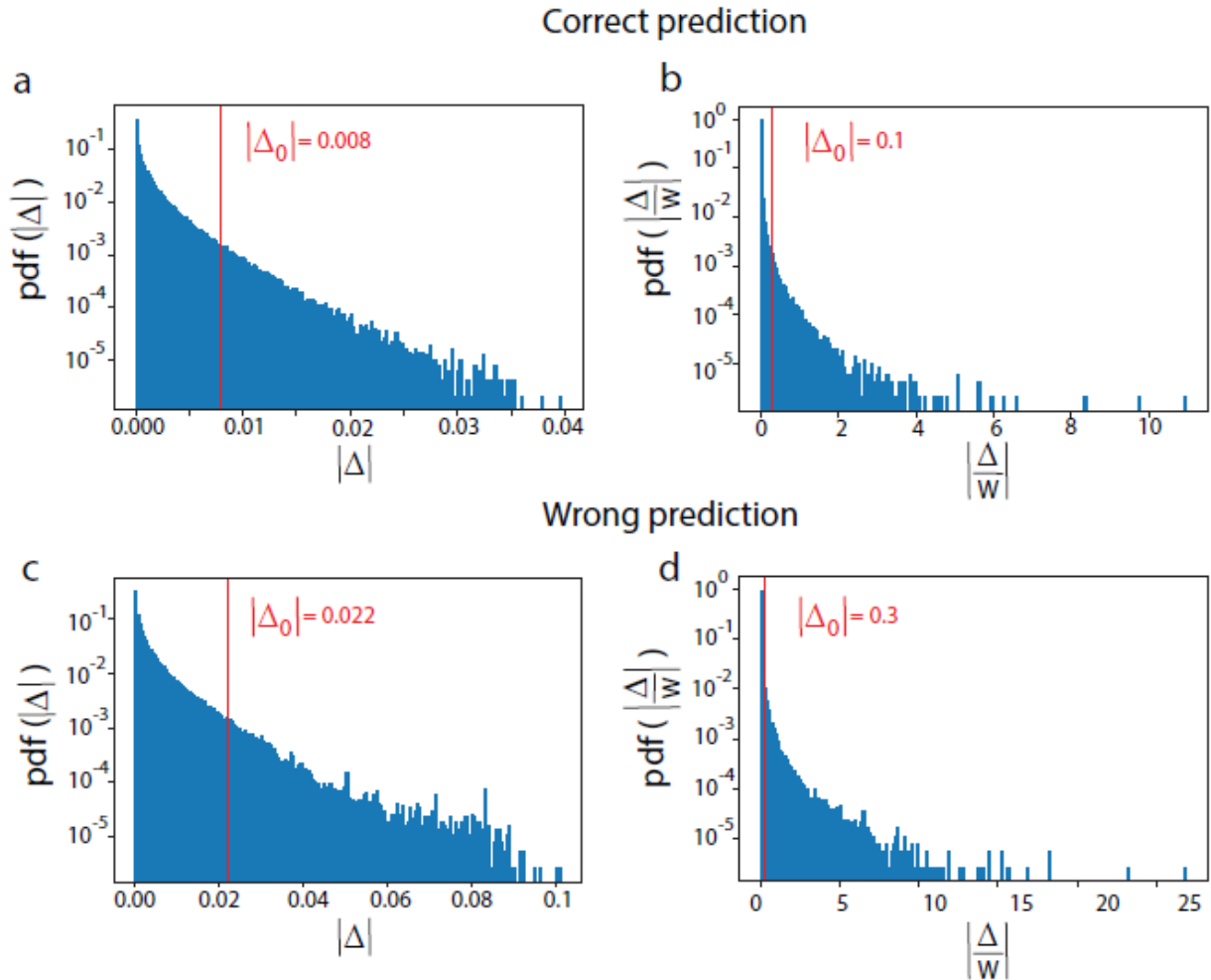

**Supplementary Figure S1. Probability density function (pdf) of $W^{Conv}$ gradients, $|\Delta|$, and $|\Delta/W^{Conv}|$ for Tree-3 using ReLU activation function. a**, pdf of the absolute value of $|\Delta|$ for a test example of with a correct predicting label. The vertical red line stands for $\Delta_0$ (denoted in red) such that the summation of pdf($\Delta < \Delta_0$) ~0.97. **b**, Similar to (**a**) for $|\Delta/W^{Conv}|$. **c**, Similar to (**a**) with a wrong predicting label. **d**, Similar to (**b**) with a wrong predicting label. Each one of the histograms consists of 1000 bins. In all panels, the vertical axis is in log-scale.

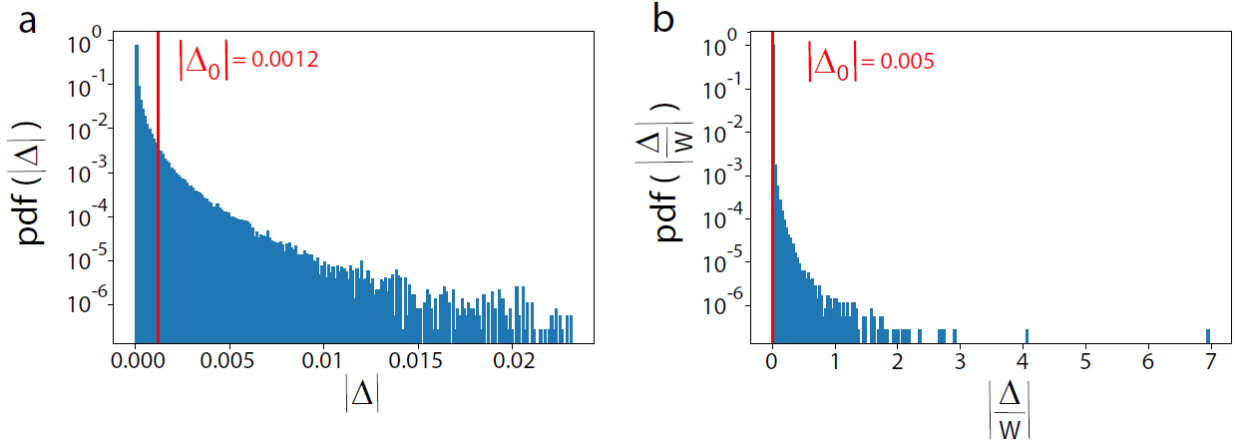
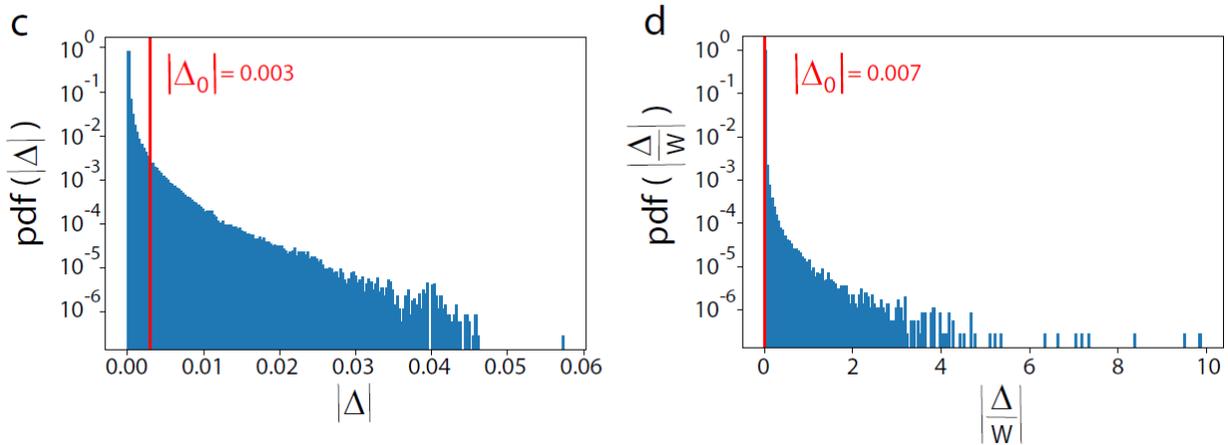

**Supplementary Figure S2. Probability density function (pdf) of $W^{Conv}$ gradients, $|\Delta|$, and $|\Delta/W^{Conv}|$ for Tree-3 using Sigmoid activation function. a**, pdf of the absolute value of $|\Delta|$ for a test example with a correct predicting label. The vertical red line stands for $\Delta_0$ (denoted in red) such that the summation of pdf($\Delta < \Delta_0$) ~0.97. **b**, Similar to (**a**) for $|\Delta/W^{Conv}|$. **c**, Similar to (**a**) with a wrong predicting label. **d**, Similar to (**b**) with a wrong predicting label. Each one of the histograms consists of 1000 bins. In all panels, the vertical axis is in log-scale.